\documentclass{article}

\PassOptionsToPackage{numbers, compress}{natbib}
\usepackage[sglblindworkshop, final]{neurips_2025}
\usepackage[utf8]{inputenc} % allow utf-8 input
\usepackage[T1]{fontenc}    % use 8-bit T1 fonts
\usepackage{hyperref}       % hyperlinks
\usepackage{url}            % simple URL typesetting
\usepackage{booktabs}       % professional-quality tables
\usepackage{amsfonts}       % blackboard math symbols
\usepackage{nicefrac}       % compact symbols for 1/2, etc.
\usepackage{microtype}      % microtypography
\usepackage{xcolor}         % colors
\usepackage{graphicx}       % load figures

\title{Principled Operator Learning in Ocean Dynamics: The Role of Temporal Structure}
\workshoptitle{Machine Learning and the Physical Sciences (ML4PS)}

\author{%
  Vahidreza Jahanmard~\thanks{Tallinn University of Technology, Department of Civil Engineering and Architecture}~~\quad Ali Ramezani-Kebrya \thanks{Integreat - Norwegian Centre for knowledge-driven machine learning; University of Oslo}~~\quad Robinson Hordoir \thanks{Institute of Marine Research; Bjerknes Centre for Climate Research} \\
  \\
  $^{\ast}$\href{mailto:vahidreza.jahanmard@taltech.ee}{\texttt{vahidreza.jahanmard@taltech.ee}}
}

\begin{document}

\maketitle

\begin{abstract}
  Neural operators are becoming the default tools to learn solutions to governing partial differential equations (PDEs) in weather and ocean forecasting applications.
  Despite early promising achievements, significant challenges remain, including long-term prediction stability and adherence to physical laws, particularly for high-frequency processes.
  In this paper, we take a step toward addressing these challenges in high-resolution ocean prediction by incorporating temporal Fourier modes, demonstrating how this modification enhances physical fidelity. This study compares the standard Fourier Neural Operator (FNO) with its variant, FNOtD, which has been modified to internalize the dispersion relation while learning the solution operator for ocean PDEs. The results demonstrate that entangling space and time in training of integral kernels enables the model to capture multiscale wave propagation and effectively learn ocean dynamics. FNOtD substantially improves long-term prediction stability and consistency with underlying physical dynamics in challenging high-frequency settings compared to the standard FNO. It also provides competitive predictive skill relative to a state-of-the-art numerical ocean model, while requiring significantly lower computational cost.

\end{abstract}

\section{Introduction}
Accurate and timely ocean forecasting is essential for climate resilience, maritime safety, and sustainable marine resource management.
Operational systems primarily rely on physics-based numerical models, which face limitations such as high computational costs, discretization requirements, and dependence on parameterized subgrid processes \citep{fox2019challenges}. These factors constrain model scalability, adaptability across geographic domains and evolving physical regimes, and the ability to perform efficient and skillful ensemble forecasting.
With climate change amplifying extreme ocean events, the demand for fast, accurate, reliable forecasts to support early warning and decision-making has become more critical than ever \citep{ipcc2019srocc}. Neural operators are increasingly explored in Earth science, particularly in ocean and weather forecasting, as scalable alternatives to computationally expensive numerical simulations, capable of learning surrogate models for complex physical systems \citep{pathak2022fourcastnet, bonev2023spherical, chattopadhyay2024oceannet, xiong2023ai}. 

Neural operators are designed to learn nonlinear mappings between infinite-dimensional function spaces, making them particularly well-suited for approximating solutions to complex systems governed by partial differential equations (PDEs) \citep{li2020fno, Kovachki2023no, azizzadenesheli2024neural}. However, as noted by \citet{bartolucci2023neural}, an important concern arises: beyond defining the operator at the continuous level, establishing a form of continuous–discrete equivalence is essential for a neural architecture to faithfully learn the underlying operator.

This paper takes a step toward addressing the operator learning challenge in high-resolution ocean dynamics through the lens of ocean wave propagation. To bridge the gap between the continuous operator and the discrete ocean model, we must adhere to the principles of wave solutions. To this end, we argue that integral kernels should be parametrized jointly over space and time, rather than solely mapping the current state to the next time step at a fixed interval. This allows the operator to internalize dispersion relations, a crucial step toward capturing the underlying physics in a principled manner. As a result, the model not only enables principled zero-shot super-resolution in operator learning by leveraging the fact that information propagates through the medium via waves, but also facilitates effective training on limited datasets, even beyond regular gridded dataset \citep{lingsch2023beyond}.
In our experiments, we conduct a comparative analysis of the standard Fourier Neural Operator (FNO) architecture and our modified version (FNOtD), with a particular focus on evaluating their physical fidelity. To do so, we train both models under the same conditions to predict hourly sea level at a regional scale and evaluate their sensitivity to perturbations in the initial conditions. 

\section{Method}\label{sec.method}
\paragraph{Problem definition}
Let $\mathcal{G}_\theta: a \mapsto	u$ be a parametric mapping based on the FNO framework to approximate the true solution operator $\mathcal{G}^\dagger$ of the PDEs defined, where $(a, u) \in \mathcal{A} \times \mathcal{U}$ \citep{li2020fno}. Here, the underlying PDEs are the shallow water equations, and we modified the mapping from a sequence of input variables $\bar{X}(x,y,t-(\tau+1)\Delta t:t)$ to a sequence of output variables $X(x,y,t+\Delta t:t+\tau\Delta t)$ where $\tau \Delta t$ represents the time window used to capture the temporal structure ($\tau = 1$ in the standard FNOs). Input variables $\bar{X}$ include initial states along with forcing inputs derived from known dynamic factors relevant to the forecasting problem. 

\paragraph{Operator architectures}
We consider two variants of the FNO architecture: the standard model, which applies Fourier transforms solely in the spatial domain ($\tau=1$), and a modified version (FNOtD), where the integral kernels $\mathcal{K}$ are parameterized jointly over space and time. The architectures are defined as follows, with the number of Fourier layers set to $L=4$:
\begin{equation}\label{eq:01}
    \mathcal{G}_\theta := Q \circ 
    \sigma_L (W_{L-1} + \mathcal{M}_{L-1}(\mathcal{K}_{L-1} + b_{L-1})) \circ \dots \circ 
    \sigma_1(W_0 + \mathcal{M}_{0}(\mathcal{K}_0 + b_0)) \circ P
\end{equation}
where $\mathcal{M}$ is a point-wise multilayer perceptron enhancing local nonlinear representation \citep{qin2024toward}. The integral kernel operators are defined as $\mathcal{K}(x) =  \mathcal{F}^{-1} \left( R \cdot \mathcal{F}(v_j)\right)(x)$, where $\mathcal{F}$ and $\mathcal{F}^{-1}$ denote the Fourier and inverse Fourier transforms, respectively, and $R$ is a complex-valued weight tensor in the Fourier domain. The key difference between the models lies in the dimensionality $R$: for standard FNO, $R$ is defined over spatial frequency modes only, while in FNOtD, it is extended to include the temporal frequency dimension, resulting in a higher-dimensional weight tensor. In both cases, the Fourier modes are truncated to $k_x^{\mathrm{max}}$, $k_y^{\mathrm{max}}$, and, additionally for FNOtD, $\omega{\mathrm{max}}$ in the temporal domain. To ensure a fair comparison, we adjusted the truncation thresholds such that both models have approximately the same number of learnable parameters. Architectural details are provided in Appendix~\ref{appendixA}.

\paragraph{Training objective}
The models are trained to minimize the relative L2 error between predictions $\hat{y}$ and target values $y$, a commonly used loss function in operator learning \citep{Kovachki2023no}. The FNOtD model enables distributing the loss computation across the predicted lead time window $t_i+(\Delta t:\tau \Delta t)$, which can facilitate learning high-frequency events from datasets with low signal-to-noise ratio. However, in this study, we compute the loss using only the first lead time step $l=1$ to ensure both models are optimized under identical supervision targets. The loss function $\ell$ is defined as the sum of individual losses for each output variable $v$, with $\ell_v$ expressed in its general form as: 
\begin{equation}\label{eq:03}
    \ell_v(l) = \frac{1}{N_s} \sum_{k=1}^{N_s} \frac{\| \hat{y}_v(l)[m,n,k] - y_v(l)[m,n,k] \|_2}{\| y_v(l)[m,n,k] \|_2} , \quad \forall (x_m, y_n) \notin M_{\mathrm{land}}
\end{equation}
where $N_s$ denotes the number of samples and $\|.\|_2$ represents the L2-norm computed over the spatial dimensions $[m,n]$. A land mask $M_{land}$ is applied to exclude land grid points from the loss computation. The scope of this study is limited to predicting sea level as the only output variable.

\paragraph{Experiment}
We evaluate both variants of the FNO architecture in a regional ocean setting, using the Baltic Sea as a case study. This shallow, non-tidal, stratified, semi-enclosed basin in Northern Europe features complex morphology and bathymetry. Its multiscale dynamics driven by atmospheric forcing, thermal stratification, and salinity gradients make it a challenging testbed for modeling multiscale ocean dynamics \citep{weisse2019baltic, jahanmard2023quantification}. Both models are trained on the same dataset for 50 epochs using the Adam optimizer with a cosine annealing learning rate schedule, starting from an initial rate of $1 \times 10^{-3}$, on a single NVIDIA A100-80GB GPU, with a training time of approximately six hours per model. Training is performed on a low-resolution version of the data, while evaluation is carried out on a high-resolution grid to generate hourly sea level predictions. The low-resolution training data is obtained by resampling the high-resolution grid using bilinear interpolation. To improve generalization and reduce dependence on specific grid alignments, we introduce randomness in the resampling step by selecting varied starting points, effectively sliding the coarse grid over the fine-resolution domain. This approach preserves the spatial extent and uniform grid spacing, enabling the learned operator to better approximate the underlying continuous dynamics independent of specific discretizations.

\paragraph{Dataset}
We train and evaluate the models using high-resolution ocean simulations from the Baltic Sea Physics Analysis and Forecast product, hereafter referred to as the Nemo model. This model provides hourly three-dimensional outputs with a horizontal resolution of one nautical mile. It incorporates data assimilation from in situ temperature and salinity profiles, satellite-derived sea surface temperature (SST), and sea ice concentration.
The input variables for both FNO variants include sea level, sea surface temperature, sea surface salinity, reference water depth, and atmospheric forcing fields (refer to Appendix \ref{appendixA}). The atmospheric forcing consists of sea level pressure and 10-meter zonal and meridional wind components, sourced from the ERA5 reanalysis dataset.
The FNO models are trained to forecast sea level, using the corresponding Nemo outputs as the target. The dataset spans from November 2021 to December 2024. The training period covers November 2021 to September 2024, and the test period spans September to December 2024, with a ten-day temporal gap in between. 
Although the training duration is relatively short, it suffices to assess the performance of the proposed FNOtD architecture, particularly in terms of physical fidelity and long-term stability, compared to the standard FNO model.

\paragraph{Inference}
The trained models are evaluated in an autoregressive inference setting, where sea level is predicted iteratively under prescribed atmospheric forcing and sea surface salinity and temperature. Each forecast is initialized with the initial condition at a specified origin time, and predictions are then rolled forward over the desired lead time horizon. Similar to numerical models, FNO models are also expected to exhibit a spin-up period to mitigate the influence of initial condition perturbations. We assess the performance of FNO models on the out-of-sample test dataset using iterative predictions.
Note that since the forcing inputs are derived from a reanalysis dataset and we aim to assess physical accuracy rather than forecasting skill in this study, we hereafter refer to the task as a prediction instead of forecasting.

\section{Results}
\begin{figure}
  \centering
  \includegraphics[width=\textwidth]{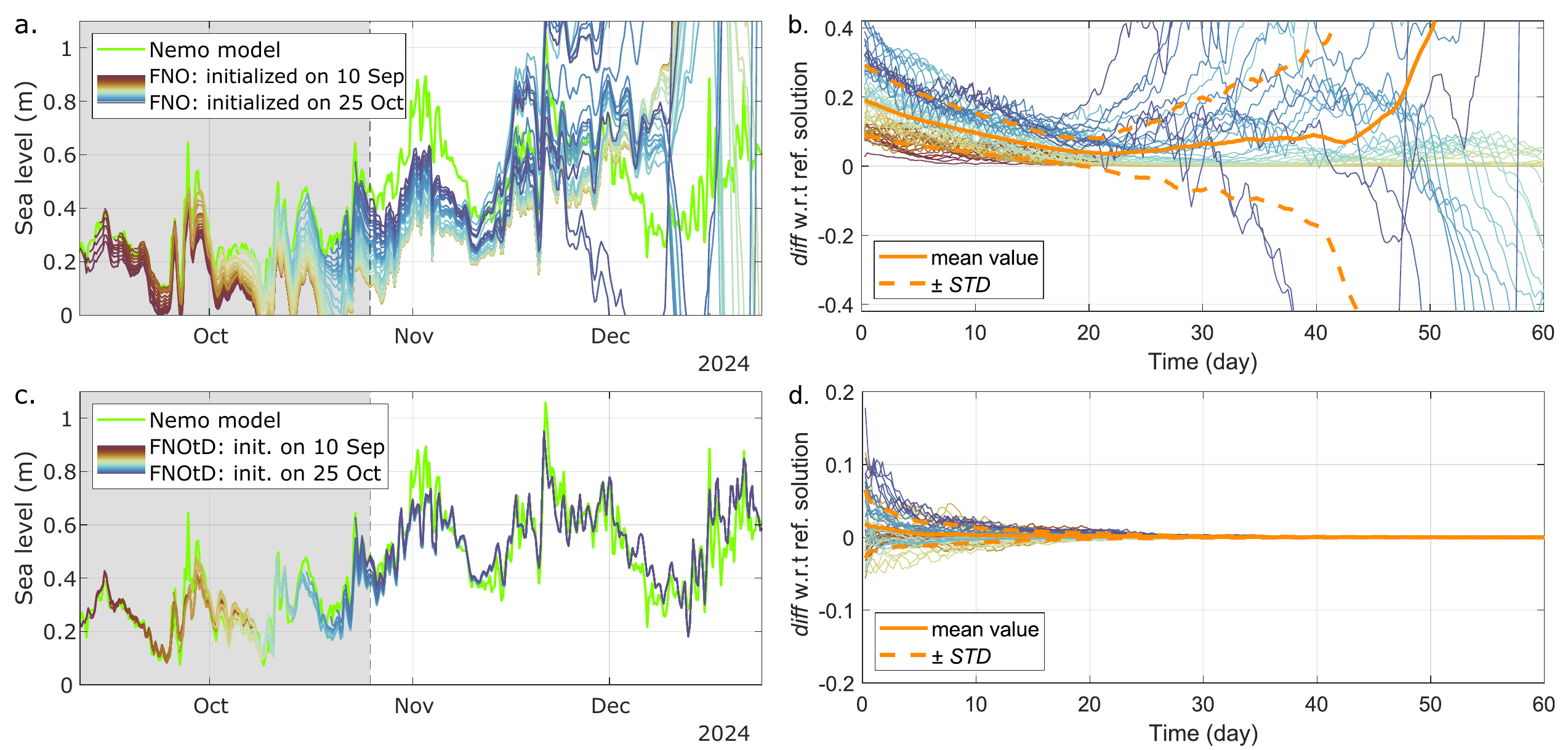}
  \caption{Iterative predictions with different initial conditions. Time series of outputs at a representative location are shown for (a) the standard FNO and (c) FNOtD. The right panels illustrate the convergence of the predictions under identical forcing by displaying the difference between each prediction and a reference prediction initialized on September 10. Their time axis is relative to each prediction’s origin time, while differences are computed at corresponding calendar timestamps. Repeating the model training with different random initializations yields consistent results.}
  \label{fig:01}
\end{figure}
\paragraph{Sensitivity to initial conditions}
While perturbations in initial conditions can significantly affect the early phases of ocean simulations, their influence diminishes over time, especially in surface layers. In ocean-only models without ocean–atmosphere feedback, surface fields are primarily driven by atmospheric forcing \cite{tokmakian2019influence}. Therefore, we leveraged this known physics of ocean simulation to illustrate the physical realism of the trained models. 
Note that although the FNO models are trained on outputs from the Nemo model, the sea level fields produced by Nemo itself exhibit RMSE of 0.08 m relative to a network of geoid-referenced tide gauge observations along the Baltic coast \cite{jahanmard2023quantification}.
Therefore, this inherent level of random noise in the Nemo model enables the sensitivity experiment by applying iterative predictions from varying origin times. Figure \ref{fig:01} illustrates this experiment for both FNO variants by rolling the initial conditions and performing iterative predictions on the test dataset. We observe that the standard FNO displays notable instability, with predictions diverging rather than converging after a few iterative steps, contrary to expectations. In contrast, FNOtD exhibits stable convergence of the predictions after approximately three weeks (Figure \ref{fig:01}d). 
This period is analogous to the spin-up phase in ocean modeling, during which the model adjusts to reach a statistically steady state under applied external forcing. A similar three week period for reconstructing ocean circulation under atmospheric forcing with a deep learning model was also observed by \citet{Hordoir2025Atmos}.
Therefore, a more robust model evaluation is conducted after the spin-up period has elapsed, during which the standard FNO exhibits divergence and lacks long-term stability. For November and December 2024, FNOtD yields an RMSE of 0.09 m (relative RMSE of 0.18) against Nemo model, evaluated 30 days after initial condition.

\begin{figure}[htbp]
  \centering
  \includegraphics[width=\textwidth]{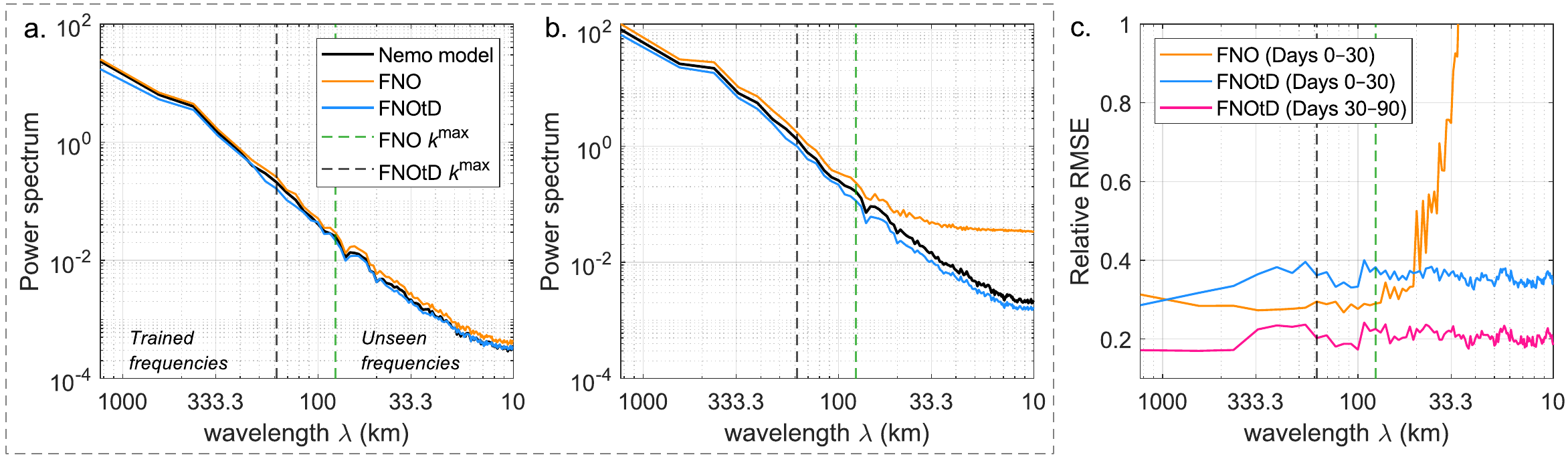}
  \caption{Radially averaged power spectra of iterative hourly predictions at (a) 7 days and (b) 30 days after prediction initialization. (c) Relative RMSE of the predictions in the spectral domain (see Appendix \ref{appendixC}). All panels correspond to forecasts initialized on October 1.}
  \label{fig:02}
\end{figure}

\paragraph{Spectral bias}
In Fig. \ref{fig:02}, the spectral bias of the models is evaluated against high-resolution Nemo outputs by calculating radially averaged power spectra. The standard FNO exhibits error propagation from high to low frequencies upon encountering the first out-of-distribution event, whereas the FNOtD demonstrates robust stability in iterative predictions. In terms of relative RMSE in the spectral domain, FNOtD outperforms the standard FNO and exhibits superior spectral stability in long-term predictions. Iterative predictions, compared with Nemo outputs and observations, are presented in Appendix \ref{appendixB} for multiple measurement stations.

\section{Limitations and Future Work}
Although the results demonstrate a promising direction for high-resolution ocean forecasting, the proposed approach has thus far been evaluated only in a single representative region. Further validation across diverse geographical settings and oceanic conditions is warranted. Future work could explore extended training data, additional predictive variables, and further model fine-tuning. Applying this framework to other marine environments or incorporating different oceanographic parameters also presents a promising research direction. One limitation of the proposed architecture, compared to the standard FNO, is the increased memory demand associated with incorporating temporal modes during training. However, the FNOtD architecture demonstrates the ability to learn and generalize from limited training data, whereas the standard FNO typically requires a larger dataset to achieve comparable generalization. Furthermore, our observations indicate that the FNOtD architecture demonstrates potential for application in training and/or fine-tuning on irregular and sparse observational datasets. In such settings, the integration of a spatiotemporal training objective with Fourier-domain parameterization of the integral kernels offers a robust framework for constraining operator learning, thereby enabling the model to effectively capture evolving underlying dynamics within limited and irregular training data.

\section{Conclusion}
We introduced a variant of the Fourier Neural Operator (FNO) framework tailored for ocean modeling, where the model is guided by the principles of ocean wave propagation. 
Our results show that integrating temporal Fourier modes enhances operator learning, leading to improved physical accuracy and long-term emulation stability.
These findings point to a promising direction for AI-enabled ocean digital twins. The proposed architecture demonstrates both physical fidelity and robustness, making it suitable for operational ocean applications. By enabling rapid, computationally efficient ensemble forecasting, it supports climate adaptation and resource management. FNO-based models can generate large ensemble forecasts at computational costs several orders of magnitude lower than state-of-the-art numerical models. Future work will extend this framework to three-dimensional emulation and incorporate additional predictive variables to further improve robustness and applicability.

\section*{Acknowledgments}
This research has been supported by the Estonian Research Council grant no. PRG1785 and the Tallinn University of Technology grant no. GFEAVJ24. Ali Ramezani-Kebrya was supported by the Research Council of Norway through FRIPRO Grant under project number 356103 and its Centres of Excellence scheme, Integreat - Norwegian Centre for knowledge-driven machine learning under project number 332645.

\bibliographystyle{unsrtnat}
\bibliography{bibtex}

\clearpage

\appendix

\section{Model architecture and training dataset} \label{appendixA}
In this work, we compare the performance of two variants of FNO architectures: standard FNO and FNOtD (where 'tD' refers to the incorporation of temporal Fourier modes). The standad FNO learns the mapping from $\bar{X}(x,y,t)$ to $X(x,y,t+1\Delta t)$, where $\bar{X}(x,y,t)$ comprises the output variables $X(x,y,t)$ alongside other predictive variables listed in Table \ref{tbl:01} at time $t$. Input and output variables are obtained form Nemo model\footnote{Obtained from Copernicus Marine Service, Baltic Sea Physics Analysis and Forecast product (version 202311), \url{doi:10.48670/moi-000103-7086abc0ea0}} , ERA5 reanalysis dataset\footnote{Obtained from Copernicus climate data store, \url{https://doi.org/10.24381/cds.adbb2d47}}, and GEBCO\_2024 dataset\footnote{Obtained from GEBCO Compilation Group (2024) GEBCO 2024 Grid, doi:10.5285/1c44ce99\-0a0d\-5f4f\-e063\-7086abc0ea0f}. A fixed time step of $\Delta t=6$ hours is used consistently throughout this study; for brevity, we omit the explicit notation where appropriate. Therefore, the integral kernel acts exclusively on the spatial dimensions via a 2D FFT with truncated modes $(k_x^{\mathrm{max}},k_y^{\mathrm{max}})$, consistent with previous works \citep{pathak2022fourcastnet, chattopadhyay2024oceannet}. FNOtD extends the parameterization of the integral kernels to learn from the time dimension as well, with mode truncation at ($k_x^{\mathrm{max}},k_y^{\mathrm{max}},\omega^{\mathrm{max}}$). Therefore, the operator learns mapping from $\bar{X}(x,y,t-\tau+1:t)$ to $X(x,y,t+1:t+\tau)$, where $\tau$ is a fixed temporal window. Our findings indicate that a temporal window of two days is suitable for learning wave propagation patterns at the regional scale. However, tuning the window size and the number of Fourier modes enables the model development to be adapted for longer-term forecasting tasks. 

\begin{table}[htbp]
  \caption{Input and output variables. For the standard FNO, the parameter $\tau$ is set to 1. All variables are projected from the original Gaussian grid onto a uniform Cartesian grid. Additionally, datasets not originating from the Nemo model are interpolated onto the Nemo grid using bilinear interpolation. }
  \label{tbl:01}
  \centering
  \begin{tabular}{ll}
    \toprule
    Input $\bar{X}(x,y,t-\tau+1:t)$ & Output $X(x,y,t+1:t+\tau)$ \\ 
    \midrule
    Sea level (SL) -- Nemo & Sea level (SL) -- Nemo \\ 
    Sea surface temperature (SST) -- Nemo \\ 
    Sea surface salinity (SSS) -- Nemo \\ 
    Sea level pressure -- ERA5 &  \\ 
    U wind -- ERA5 & \\ 
    V wind -- ERA5 & \\ 
    Reference water depth -- GEBCO  & \\ 
    \bottomrule
  \end{tabular}
\end{table}

\begin{figure}
  \centering
  \includegraphics[width=.8\textwidth]{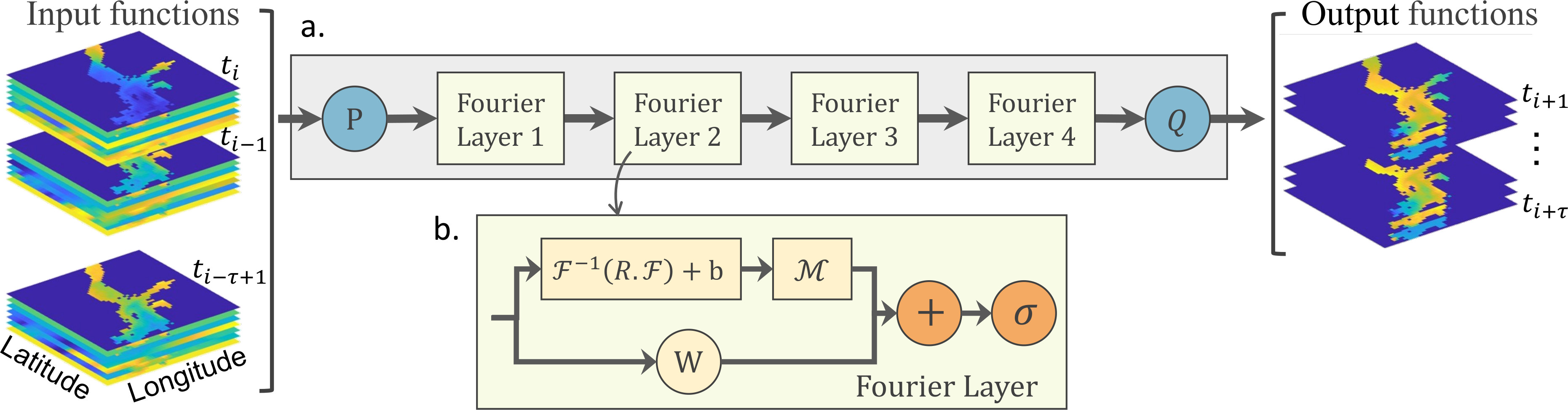}
  \caption{(a) Architecture of the Fourier Neural Operator FNOtD, with input and output dimensions of h×w×t×c, where h and w correspond to the latitude and longitude grid dimensions, t denotes time steps, and c is the number of physical variables. Eliminating the temporal dimension yields the standard FNO model architecture. The input is first projected into a higher-dimensional channel space by a point-wise encoder P. After passing through four Fourier layers, the feature representations are mapped to the output variables via a point-wise decoder Q. (b) Schematic of a single Fourier layer. $\mathcal{M}$ is a point-wise multilayer perceptron layer to enhance local nonlinear representation learning.}
  \label{fig:03}
\end{figure}

This modification enables the operator to treat the time axis as grid-independent, allowing the model to capture multiscale variations by learning the dispersion relation. In the standard FNO, the time step is fixed during training, which makes it difficult for the model to learn the mapping multiscale processes and extrapolate to unseen frequencies. Figure \ref{fig:03} shows the FNOtD architecture used in this study, with four Fourier layers. The standard FNO architecture will be obtained from this figure by setting $\tau=1$. In the standard FNO, the truncated Fourier modes are set as $k_x^{\mathrm{max}} = k_y^{\mathrm{max}} = 16$, whereas in FNOtD, the modes are set to $k_x^{\mathrm{max}} = k_y^{\mathrm{max}} = 8$ and $\omega^{\mathrm{max}} = 4$. This setup ensures a roughly equivalent number of learnable parameters in both models (FNO: 16.71 M; FNOtD: 16.19 M). However, in our experiments, the standard FNO with 8 modes demonstrates comparable performance to that with 16 modes.

Note that in FNOtD, the first lead time ($l=1$, as defined in Equation \ref{eq:03}) is used for loss computation to ensure consistency with the standard FNO. However, this architecture allows for loss computation across multiple lead times, thereby facilitating efficient constraint of operator learning in chaotic dynamical systems with measurement uncertainty. Designing principled neural operators enable effective learning from physical laws, even in the absence of extensive or regularly sampled datasets. The results from the FNOtD architecture show promising direction for training or fine-tuning models directly on sparse (in time and space) observational data, such as in-situ or satellite observations.

In this paper, we demonstrate that FNOtD can efficiently capture multiscale processes, enable principled extrapolation to unseen frequencies, and generalize effectively beyond the training dataset. These capabilities address key challenges associated with the standard FNO, which mostly improved by extensive training in previous studies. The spatiotemporal dependencies in FNOtD also allow for efficient fine-tuning, even in cases with limited data or when loss functions include masked or missing observations.
Nevertheless, incorporating the time dimension in FNOtD increases the data load during training, leading to higher GPU memory requirements and requiring additional effort to optimize the training setup relative to the standard FNO. Furthermore, FNOtD is more prone to training plateaus, particularly when the dataset contains high levels of noise that obscure underlying physical relationships, such as dispersion relations or when the input variables fail to adequately represent the governing dynamics.
While FNOtD offers a promising principled framework for operator learning in ocean dynamics, it has so far been evaluated in a single representative region. Therefore, its generalizability across different regions and scenarios remains to be validated across diverse geographical settings and under varying conditions in future work.

\section{Iterative predictions against Nemo output and observations} \label{appendixB}
\begin{figure}
  \centering
  \includegraphics[width=\textwidth]{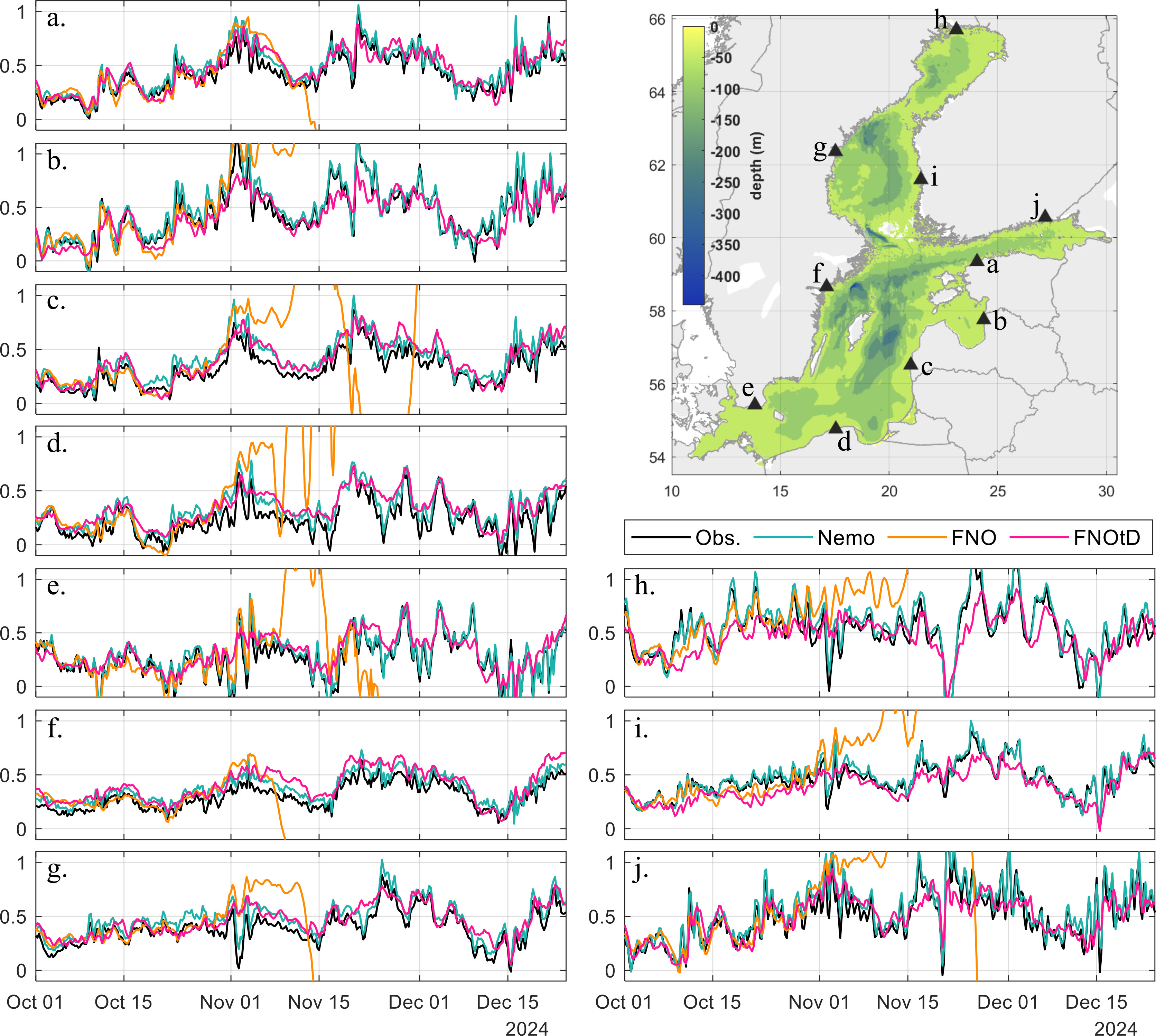}
  \caption{Time series of iterative sea level (in meter) predictions by the standard FNO and FNOtD are shown, alongside Nemo outputs at the nearest grid point to the corresponding tide gauge observation. The data corresponds to the period of the out-of-sample test dataset.}
  \label{fig:04}
\end{figure}

Figure \ref{fig:04} presents the iterative predictions of the standard FNO and FNOtD, with initial conditions set on October 1, for multiple measurement stations along the Baltic coastline. The predictions are compared with Nemo outputs and geoid-referenced tide gauge records \citep{jahanmard2023quantification}. In this figure, it is evident that the standard FNO model (represented by the orange lines) exhibits significant instability across all stations when encountering the first out-of-distribution event, leading to a dramatic degradation in performance. As shown in Figure \ref{fig:01}, this instability increases as the model moves further from the training period and encounters additional out-of-distribution samples. The instability arises from error propagation across high to low frequencies, highlighting the difficulty of the standard FNO in learning the underlying operator that maps multiscale processes. 

In contrast, FNOtD (represented by the magenta lines) demonstrates stable predictions, which are comparable to the Nemo outputs, the state-of-the-art numerical high-resolution 3D ocean model for the Baltic Sea (green lines). The RMSE of FNOtD and Nemo outputs relative to tide gauge observations are 0.12 m and 0.08 m, respectively. Note that the FNOtD was trained on a limited dataset, allowing for a comparison with the equivalent standard FNO.
Hence, a larger model trained on a more extensive reanalysis dataset and involving more predictive variables, potentially incorporating variables from multiple depths, could improve prediction accuracy and enhance the model's ability to capture extreme events with better uncertainty quantification. This accuracy can be further improved through fine-tuning on observational data. Such capabilities pave the way for data-driven ocean forecasting with predictive skill comparable to, and potentially surpassing, that of state-of-the-art numerical models, while also enabling ensemble predictions at orders of magnitude lower computational cost.

\section{Evaluation metrics} \label{appendixC}
The root mean squared error (RMSE) in its general form is defined as:
\begin{equation}\label{eq:A1}
    \mathrm{RMSE} = \sqrt{ \frac{1}{MNQ} \sum_{m,n,q} \left( \hat{y}[m,n,q] - y[m,n,q] \right)^2 }, \quad \forall (x_m, y_n) \notin M_{\mathrm{land}}
\end{equation}
where $\hat{y}$ and $y$ are predicted and reference values at the location ($x_m,y_n$) and prediction time $t_q$, and $M$, $N$, and $Q$ denote the total number of samples in the x and y directions, and time, respectively. The land mask ($M_{land}$) is applied to exclude land grid points. In the RMSE computation with respect to observations, $x_m$ and $y_n$ represent the spatial coordinates of the observation stations.

The Relative RMSE (RRMSE) of the radially averaged power spectrum shown in Figure \ref{fig:02}c is determined at wavenumber $k$ (or equivalently, wavelength $\lambda$, or Fourier modes), as follows:

\begin{equation}\label{eq:A5}
    \mathrm{RRMSE}(k) = \sqrt{ \frac{\sum_{q} \left( P_{\hat{y}}(k)[q] - P_y(k)[q] \right)^2}{\sum_{q} \left( P_y(k)[q] \right)^2}}
\end{equation}
where $P_{\hat{y}}$ and $P_y$ denote the radially averaged power spectra of the predicted and referenced values, respectively. The radially averaged power spectrum at wavenumber $k$ is the average of a 2D power spectrum $PSD$ whose radial frequency $\sqrt{(k_x^2+k_y^2 )}$ falls within a small interval around $k$:

\begin{equation}\label{eq:A6}
    P_v(k) = \langle PSD(k_x,k_y) \rangle_{k_x^2+k_y^2 \approx k}
\end{equation}
where $\langle . \rangle$ denotes the average operator.

\end{document}